\documentclass[journal]{IEEEtran}

\usepackage[utf8]{inputenc}
\IEEEoverridecommandlockouts                              

\usepackage{graphicx}
\usepackage{caption}
\usepackage{subcaption}
\usepackage{array,multirow}
\usepackage{makecell}
\usepackage{booktabs}
\usepackage{booktabs}
\usepackage{float}
\usepackage{hyperref}
\usepackage{url}
\usepackage{pifont}
\newcommand{\xmark}{\ding{55}}%
\usepackage{tabularx}
\usepackage{adjustbox}
\usepackage[dvipsnames]{xcolor}
\usepackage{color}
\hypersetup{
    colorlinks=true,
    linkcolor=blue,
    citecolor=blue,      
    urlcolor=blue,
}
\usepackage{mathptmx} 
\usepackage{amsmath} 
\usepackage{soul}

\title{\Huge \bf Uncertainty-Aware Credit Card Fraud Detection Using Deep Learning
}

\author{Maryam Habibpour, Hassan Gharoun, Mohammadreza Mehdipour, AmirReza Tajally, Hamzeh Asgharnezhad \\ Afshar Shamsi, Abbas Khosravi,~\IEEEmembership{Senior Member,~IEEE}, Miadreza Shaﬁe-khah,~\IEEEmembership{Senior Member,~IEEE}\\ 
Saeid Nahavandi,~\IEEEmembership{Fellow,~IEEE}, and João P.S. Catalão,~\IEEEmembership{Senior Member,~IEEE}

\thanks{This research was partially supported by the Australian Research Council's Discovery Projects funding scheme (project DP190102181).}
\thanks{M. Habibpour is with Lian Sazeh group, Tehran, Iran (e-mail: E.maryamhabibpour@gmail.com)}
\thanks{H. Gharoun is with the School of Industrial Engineering, College of Engineering, University of Tehran, Tehran, Iran (e-mail: h.gharoun@alumni.ut.ac.ir)}
\thanks{A. Tajally is with the Department of Industrial Engineering, university of Tehran, Tehran, Iran (e-mail: Amirreza73tajally@gmail.com)}
\thanks{M. Mehdipour, H. Asgharnezhad, and A. Shamsi are individual researchers, Tehran, Iran (e-mail: ishahaab@gmail.com, Hamzeh.asgharnezhad@gmail.com, afshar.shamsi.j@gmail.com)}

\thanks{A. Khosravi,  and S. Nahavandi are with the Institute for Intelligent Systems Research and Innovation (IISRI), Deakin University, Australia (e-mail: \{abbas.khosravi, saeid.nahavandig\}@deakin.edu.au)}

\thanks{M. Shaﬁe-khah is with the School of Technology and Innovations, University of Vaasa, 65200 Vaasa, Finland.}
\thanks{J.P.S. Catalão is with the Faculty of Engineering of University of Porto and INESC TEC, Porto, Portugal.}
}

\begin{document}

\maketitle
\thispagestyle{empty}
\pagestyle{empty}

\begin{abstract}

Countless research works of deep neural networks (DNNs) in the task of credit card fraud detection have focused on improving the accuracy of point predictions and mitigating unwanted biases by building different network architectures or learning models. Quantifying uncertainty accompanied by point estimation is essential because it mitigates model unfairness and permits practitioners to develop trustworthy systems which abstain from suboptimal decisions due to low confidence. Explicitly, assessing uncertainties associated with DNNs predictions is critical in real-world card fraud detection settings for characteristic reasons, including (a) fraudsters constantly change their strategies, and accordingly, DNNs encounter observations that are not generated by the same process as the training distribution, (b) owing to the time-consuming process, very few transactions are timely checked by professional experts to update DNNs. Therefore, this study proposes three uncertainty quantification (UQ) techniques named Monte Carlo dropout, ensemble, and ensemble Monte Carlo dropout for card fraud detection applied on transaction data. Moreover, to evaluate the predictive uncertainty estimates, UQ confusion matrix and several performance metrics are utilized. Through experimental results, we show that the ensemble is more effective in capturing uncertainty corresponding to generated predictions. Additionally, we demonstrate that the proposed UQ methods provide extra insight to the point predictions, leading to elevate the fraud prevention process.

\end{abstract}

\section{INTRODUCTION}
\label{sec:I}
One of the main pitfalls related to modern e-commerce is the increase in card fraud. There is public concern about card-present (CP) and card-not-present (CNP) transactions. Although banks have developed chip smart cards leading to a significant drop in CP fraud, the main issue is online payments (CNP). Modern encryptions and state-of-the-art multi-factor authentication (MFA) techniques such as biometric technology have been developed to disrupt fraudulent activities, preserve the reputation of card issuers and merchants, and therefore diminish monetary losses for customers \cite{consumer}. Nevertheless, perpetrators of fraud always find a blind spot to exploit. 

In recent circumstances, the emergence of the coronavirus (COVID-19) pandemic has accelerated the shift towards a more digital world and likewise expanded the e-commerce market. This widespread disease has exacerbated the circumstances and has led fraudsters to target individuals' financial assets to a considerable extent. According to the statistics released by the federal trade commission (FTC), over \$3.3 billion fraud losses were reported in the 2020 epidemic, as compared with \$1.8 billion in 2019; almost \$1.2 billion of losses were attributed to imposter scams, while about \$246 million of losses were the consequences of online shopping \cite{federaltradecommission_2021}. 

Due to the expansion of artificial intelligence (AI) technologies in various dimensions of human life, we have witnessed massive investments into AI implementation by digital security companies to identify illegitimate transactions and make barriers to stop them. Owing to the promising human-level performance, it initially appears desirable to implement deep neural networks (DNNs) to detect card fraud. However, these efficiencies are often accompanied by increased complexity, and thus, raising concerns about the reliability of modern networks. Excessive use of these black-box predictors for critical topics such as medical diagnosis \cite{shamsi2021uncertainty, khaledyan2021confidence}, and particularly this case (fraud detection) \cite{shenvi2019credit,raghavan2019fraud,carrasco2020evaluation}, has made it extremely necessary to draw a line between well-grounded predictions and those with challenging inference. The lack of interpretability of the mentioned complex models is by no means the only issue. The argument that is often stated is the lack of explicit information about the internal processing \cite{alain2016understanding}. These autonomous models require massive amounts of training samples to reach the optimal decision. The limited number of samples also raises questions about epistemic uncertainty \cite{postels2019sampling}. Notably, the most persuasive argument is that these models with arbitrary depth tend to have high confidence in their predictions \cite{guo2017calibration}. That is to say, in dynamic and constantly evolving environments, where customers' shopping patterns alter periodically or seasonally, nimble fraudsters employ new attacks in response to these changes. Upon such occurrences, the transformation of statistical properties occurs, and discriminative neural networks encounter observations that are not generated by the same process as the training distribution (out of distribution) \cite{liu2020simple}. Regrettably, under this circumstance, DNNs have little or no performance guarantees to generate predictions with a low confidence level. Explicitly, overconfident erroneous predictions can be detrimental \cite{amodei2016concrete}.

Besides the drawbacks of non-linear neural networks, another paramount concern appears with the existence of verification latency \cite{krempl2011classification}, in which the process of determining the status of new transactions that update the classifier is delayed. Generally, besides the terminal, which performs the primary and basic security checks on all payments in the shortest possible time \cite{van2015apate}, each transaction is scanned through two automated tools to authorize \cite{dal2017credit}. First, rule-based systems are employed to alert against fraudulent transactions; second, classifiers such as neural networks are exploited on a labeled data to inspect the nature of transactions \cite{dal2017credit}. Transactions flagged by automated tools are inspected by professional experts and the resulting feedback will feed and update the classifiers. This process is time-consuming, and only a limited number of alarms are monitored daily \cite{krivko2010hybrid}. Although there is a considerable lag between the suspicious transaction verification process and the classifiers update, the updated classifiers (here DNNs) are not yet guaranteed to generate accurate alerts, and due to the opaque nature, investigators cannot comprehend where the problem is. Thus, the alerts may be ignored or examined reluctantly. Overall, we have enough evidence to conclude that fraudulent transactions are considered genuine if they are neither inspected by investigators nor announced by customers.

Motivated by these shortcomings, and irrespective of the fact that both broad and in-depth studies have been conducted to illuminate how the machine learning (ML) models work \cite{bhatt2020explainable}, it is imperative to implement uncertainty quantification (UQ) techniques for studying DNN models behavior \cite{thiagarajan2019understanding}. The fact remains that transparency as an integral attribute of interpretable, comprehensible, and explainable AI, has become a high priority and an area of focus by several large technology corporations \cite{bhatt2020explainable}. UQ techniques are a form of transparency that can effectively manage this mission. In fact, by assessing the uncertainty corresponding to generated predictions, this type of methodology computes meaningful confidence values to specify model reliability and to support interpretability.  

The transaction prediction model needs to accommodate all the components so that the predicted feature vectors cross a threshold before the decision is made. Indeed, the uncertainty results from a lack of training data in certain areas of the input domain. When the model encounters a novel observation, it attempts to communicate with its knowledge to know it, leading to an uncertainty report appearing. The reportage of uncertainty of the model associated with the incorrect prediction could be a step for a requisite investigation and an obstacle to potential losses. Hence, investigators should consider the uncertainty prediction, not necessarily the prediction itself.

In addition to measuring the uncertainty of the generated predictions (either genuine or fraudulent), UQ techniques have recognized the fact that occasionally the inefficiency of the model is quite probable, which could be specified and minimized through some attempts (e.g., the transaction dataset is a highly skewed dataset since the genuine class outnumbers the fraudulent class, leading to higher uncertainty \cite{khan2019striking}). According to the epistemic uncertainty approach (model variability) \cite{kendall2017uncertainties}, if a model is trained by examining only a limited sample of fraudulent transactions, it will have an unacceptable performance in generating reliable predictions on that. On the other hand, it will yield fewer errors and more reliable predictions regarding legitimate samples. This issue can be addressed by increasing the number of fraudulent transactions while adopting a comprehensive approach. 

The main contributions of our task are summarized as follows. First of all, we deploy the Monte Carlo dropout (MCD), ensemble, and ensemble Monte Carlo dropout (EMCD) methods to estimate uncertainty for the transaction data that are available to the public. These are based on the frequentist approach and need to adjust the parameters slightly. Notably, there is no inherent mechanism in the model to corroborate that needs uncertainty estimation. The leading factor that necessitates the use of UQ techniques in DNNs is the calibration metric which estimates the interval between the probabilities of the classes extracted by the model and the true probabilities pertinent to those classes. Thus, by examining the errors of the proposed DNNs regarding the generated predictions, we illustrate reliable diagrams which serve as clues to validate our study. Importantly, given the UQ confusion matrix and several performance metrics, we evaluate the predictive uncertainty estimates and compare all the mentioned techniques. After obtaining the empirical outcomes, we conclude the study by drafting a statistical and scientific document.

The remainder of this paper is structured as follows. A review of multiple studies conducted in the task of card fraud detection is provided in section \ref{sec:II}. Section \ref{sec:III} provides detailed information on UQ techniques. The UQ confusion matrix and several performance metrics are introduced in section \ref{sec:IV}. The dataset and experiments are explained in section \ref{sec:V}. The obtained results are reported in section \ref{sec:VI}, and the conclusion and future research are presented in \ref{sec:VII}.

\section{Related works}
\label{sec:II}
Due to the nature of the problem, there are a few public transaction datasets, as well as, pertinent research works. Furthermore, irrespective of the differences in size, complexity, and attributes of the available data, there is a drastic discrepancy in the proportion of classes owing to the rise of genuine transactions compared with fraudulent ones. To illustrate, research \cite{shenvi2019credit} has revealed that the fully connected neural network is an efficient classifier to discern malicious transactions. Two sampling mechanisms were deployed to deal with the class imbalance: (i) under-sampling, which is reducing genuine transaction samples and (ii) over-sampling which is increasing fraudulent transaction samples. The obtained accuracies are 99.72\% and 99.67\%, respectively. Additionally, other alternative options have been propounded to address the problem of transaction ratio, such as (a) leveraging the interpolation technique instead of over-sampling, (b) synthesizing generative adversarial networks (GANs) based on their performance to inflate fraudulent observations artificially, or (c) wielding evaluation metrics other than accuracy in this area such as precision, recall, area under the receiver operating characteristic (ROC) curve (AUC), Matthews correlation coefficient (MCC), and cost of failure \cite{shenvi2019credit,raghavan2019fraud,banerjee2018comparative}. As discussed in Section \ref{sec:I}, the uncertainty is higher for classes with less representation, and this issue should be resolved \cite{khan2019striking}. 

In a previous study, criticism was made of this misconception that all types of transactions are at disposal during the initial model training, and accordingly, a basis was formulated to compare a once-trained model with an ensemble of classifiers that is consecutively updated \cite{dal2017credit}. Invoking intrusion detection techniques and deep learning methods (multi-layer perceptron (MLP), convolutional neural network (CNN), and deep auto-encoder (DAE)) was attempted to curtail the costs and burden of fraud investigations at related organizations through reducing false positives \cite{carrasco2020evaluation}. False positives refer to legal transactions that were incorrectly classified as fraudulent. Following encoding the feature vectors in a particular procedure and adjusting parameters by grid search, the optimal configuration out of the MLP was obtained with 35.16\% fewer alerts while catching 91.79\% of all the fraud observations \cite{carrasco2020evaluation}.


In \cite{pumsirirat2018credit}, DAE and restricted boltzmann machines (RBMs) as unsupervised deep learning (DL) methods were employed to find anomalies using reconstructed legitimate transactions. Another method for detecting anomalies is peer group analysis, which categorizes customers in accordance with their profiles and finds card fraud as it is outside the scope of customer's behavior \cite{weston2008plastic}. It should be noted that feature extraction using the mentioned methods fails to resolve the problems explained in section \ref{sec:I}. These features are overconfidently processed in passing the softmax layer. 

In the field of card fraud detection, many studies debated whether supervised methods yield more reliable results as opposed to unsupervised ones. As reported in \cite{niu2019comparison}, researchers considered both approaches and evaluated the proposed models' performances using the area under the receiver operating curves (AUROC) metric. Experiments revealed that extreme gradient boosting (XGB) and random forest (RF) outperform restricted boltzmann machine (RBM) and generative adversarial networks (GANs) in the range of 98.9\% and 98.8\%, respectively. 

As practical methods for managing global commercial, association rules provided insight into the normal behavior exerted in fraudulent transactions, which was used to detect and prevent credit card fraud \cite{sanchez2009association}. Modified Fisher discriminant analysis (MFDA) was used for the first time in this field. It was shown that by appending the weighted average during computation, the linear classifier concentrates on high-value transactions and discriminates them accurately, thereby avoiding weighty and significant losses \cite{mahmoudi2015detecting}. The hybrid methods based on AdaBoost and majority voting methods were considered in card fraud detection, and the noise was added to the samples to evaluate the robustness of the algorithms as well \cite{randhawa2018credit}. 

All of these studies report promising results, but most of them fail in real-world applications (as is the case typically with DNNs), where we cannot expect the distribution of observations to be the same as that of the training samples. Measuring the uncertainty of generated predictions is necessary to improve model adoption and user experience. Table \ref{tab:taxonomy} presents the contributions of the present study compared to the relevant studies in the literature.

\begin{table*}
\centering
\caption{Survey of recent studies on fraud detection}
\resizebox{\linewidth}{!}{%
\begin{tabular}{>{\centering\hspace{0pt}}m{0.029\linewidth}>{\centering\hspace{0pt}}m{0.056\linewidth}>{\centering\hspace{0pt}}m{0.079\linewidth}>{\centering\hspace{0pt}}m{0.163\linewidth}>{\centering\hspace{0pt}}m{0.236\linewidth}>{\centering\hspace{0pt}}m{0.267\linewidth}>{\centering\arraybackslash\hspace{0pt}}m{0.12\linewidth}} 
\toprule
Study      & Issue year & Fraudulent activity                   & Technique investigated                                                        & Data set                                                                                                        & Main objective                                                                                                                     & Uncertainty Quantification                                        \\ 
\toprule
\cite{shenvi2019credit}              & 2019                & Credit Card Transactions              & Fully connected neural network, sampling techniques                                                                           & 
 European credit card dataset                               & Improving point estimation with regard to class imbalances                                                                         & \xmark                                                                 \\ 
\hline
\cite{carrasco2020evaluation}             & 2020                & Credit Card Transactions              & 
MLP, CNN, DAE                                                                          & Spanish payments processing organization                                                                        & Reducing false positives                                                                                                           & \xmark                                                                 \\ 
\hline
\cite{kazemi2017using}             & 2017                & Credit Card Transactions              & OAE, SAE                                                                          & German credit card dataset                                                                                      & Improving point estimation and mitigating variances                                                                                & \xmark                                                                 \\ 
\hline
\cite{jurgovsky2018sequence}            & 2018                & Credit Card Transactions              & LSTM                                                                          & The data was collected from March to May 2015 in the corresponding organization                                 & Modeling the credit card fraud as a sequence classification problem and improving point estimation                                    & \xmark                                                                 \\ 
\hline
\cite{niu2019comparison}             & 2019                & Credit Card Transactions              & LR, KNN, SVM, DT, RF, XGB, OCSVM, DAE, RBM, GAN                                                & European credit card dataset                               & Comparing supervised and unsupervised procedures in the realm of card fraud detection                                              & \xmark                                                                 \\ 
\hline
\cite{rtayli2020enhanced}  & 2020                & Credit Card Transactions              & Vector Machine-Recursive Feature Elimination, SMOTE, RF                       & European credit card dataset, Africa PaySim mobile money transactions                  & A proposed hybrid model that improves class prediction & \xmark                                                                 \\ 
\hline
\cite{taha2020intelligent}  & 2020                & Credit Card Transactions              & LightGBM                                                                      & E-commerce transactions of 2009 UCSD-FICO Data Mining Contest, European credit card dataset & Tuning parameters of the LightGBM using a Bayesian-based algorithm to improve point estimation                                                              & \xmark                                                                 \\ 
\hline
\cite{forough2021ensemble} & 2021                & Credit Card Transactions              & LSTM, GRU                                                                     & European credit card dataset, Brazilian Bank data collected in 2004                                  & 
An ensemble model based on sequential modeling of data using deep recurrent neural networks to improve point estimation                                                                & \xmark                                                                 \\ 
\hline
\cite{itri2019performance}   & 2016                & Automobile Insurance                        & RF, Adaboost, MLP, Decision Table, SVM, Naïve Bayes, PART, J48, Logistic, SGD & Automobile Insurance Company claims                                                                             & Comparing various supervised learning algorithms in the task of fraud detection                                                                 & \xmark                                                                 \\ 
\hline
\cite{wang2018leveraging}   & 2018                & Automobile Insurance                       & Latent Dirichlet Allocation, Fully connected neural network                                             & Automobile Insurance Company claims                                                                             & Training DNN model using extracted text features of claims descriptions to detect fraud                                               & \xmark                                                                \\ 
\hline
\cite{yan2020improved}   & 2020                & Automobile Insurance                       & Adaptive genetic algorithm, Neural network                                                   & Automobile Insurance Company claims                                                                             & Combining an adaptive genetic algorithm with a neural network to improve ANNs' convergence speed and prediction accuracy in fraud detection      & \xmark                                                                 \\ 
\hline
\cite{sowah2019decision}          & 2019                & Health Insurance                      & Genetic Support Vector Machines (GSVM)                                        & Ghana National Health Insurance claims data                                                                     & A comparative study, Proposing GSVM and compare its performance with SVM kernel classifiers                                        & \xmark                                                                 \\ 
\hline
\cite{hancock2021gradient}         & 2021                & Health Insurance                      & CatBoost, LightGBM~                                                           & Medicare insurance claims                                                                                       & Analyzing Boosting algorithms in presence of categorical features                                                                  & \xmark                                                                 \\ 
\hline
\cite{gupta2021comparative}         & 2021                & Health Insurance                      & SMOTE, ADASYN,TGANs, DT, RF, XGBoost, LightGBM,  GBM                          & India health insurance data                                                                                     & Analyzing the performance of various over-sampling techniques to balance data and multiple supervised models to detect fraud                            & \xmark                                                                 \\ 
\hline
Present Study  & -                   & Credit Card Transactions & ensemble, MCD, EMCD                                                                          & Financial organization providing e-commerce payment solutions                                                   & Building a trustworthy AI system to compute confidence values of predicted outcomes in addition to point estimation                                                                         & Proposed three different techniques of quantifying uncertainties  \\
\bottomrule
\end{tabular}
}
\label{tab:taxonomy}
\end{table*}

\section{Methods to quantify uncertainty}
\label{sec:III}
It is an inescapable fact that Bayesian models have shown a systematic, broadly accepted strategy to flagging uncertainty for the long term, howbeit this concept somewhat seems to be costly \cite{lakshminarayanan2016simple}. Bayesian methods based on Bayesian probability theory \cite{bernardo2009bayesian} define a hypothesis space of plausible models a priori (before any data is observed) and exploit deductive reasoning to update these priors as more data becomes available. For example, a Gaussian process which is a Bayesian technique defines a prior distribution over a space of functions $F$ \cite{rasmussen2003gaussian}. This distribution is updated as more data becomes available, i.e., all functions that are consistent with the labels are maintained. At inference time, each of the functions yields an output, and expectation is calculated to generate the final prediction. The variance of these outputs represents a measure of uncertainty. Formulating a deep Gaussian process or other Bayesian models is relatively straightforward, but inference and computing in these models are not easy, and it is indeed problematic to determine the exact posterior distribution. Since this is the case, frequentist approach (Non-Bayesian) has become a preferred way for ML practitioners to estimate uncertainty. More generally, the proposed methods for estimating uncertainty follow two key approaches: Bayesian approach, where the distributions over model parameters represent the uncertainty \cite{blundell2015weight}, and Non-Bayesian approach, where the distributions over model outputs represent the uncertainty \cite{kong2020sde}. We examine the second approach in follow-up work and explain it in three procedures.

\subsection{Monte Carlo dropout (MCD)}
DL tools can be cast as Bayesian models to reduce the computational burden without requiring changes in optimization \cite{gal2016dropout}. The dropout method is one of the most broadly applicable tools to prevent overfitting in DNNs \cite{srivastava2014dropout}. It involves dropping hidden layer neurons at random during model training so that with each training step, a variant subset of the network architecture is measured and regulated. Finally, a sub-network is sampled from the full model. The MCD method proposed by \cite{gal2016dropout} takes this randomization technique one step further. It was shown that by using the dropout method the testing time, in addition to the training time, is approximately the same as the Bayesian approximation. In this manner, if one test data point is consecutively passed to the network, the probabilities extracted from the softmax function, which is the most popular objective function for classification, will construct an empirical distribution. In fact, for each test data point, $N$ model configurations are applied to obtain a set of outputs \{${\hat{y}}$\}. An empirical distribution is drawn over the obtained outputs, allowing the calculation of the mean value (posterior) and the measurement of uncertainty in terms of distributional variance. The predictive mean $(\mu_{pred})$ of the model for one test data point over MC iterations is evaluated as follows (\ref{eq:1}):

\begin{equation}
{\mu_{pred}}\approx \frac{1}{T} \sum_{t}^{} p(y=c\vert x,\hat{w_t}) 
\label{eq:1}
\end{equation}

\noindent where $x$ denotes the test input; by knowing $x$, $p(y=c\vert x,\hat{w_t}$  denotes the probability of y belonging to the $c$ (the output of softmax), and $\hat{w_t}$ denotes the network weight matrices which are calculated on the  $t^{th}$  iteration; $T$ denotes the number of MC iterations. The predictive entropy ($PE$) (\ref{eq:2}) can be used as the uncertainty estimate to determine the distribution variance (predictive uncertainty) \cite{gal2016dropout}:

\begin{equation}
{PE = - \sum_{c}^{} \mu_{pred}\log \mu_{pred}
\label{eq:2}}
\end{equation}

\noindent where $c$ refers to both classes. A model is more likely to be confident regarding its prediction if its $PE$ is small.

\subsection{Ensemble}
It is possible to generate ensembles of neural networks in different ways to measure uncertainty. Here we first provide an overview of ensemble learning paradigms and then describe uncertainty computation using an ensemble of deep networks. One of the most popular ensembles is random forest (RF), a subset of decision trees that encourages the training of base learners without any interaction with other base members. In contrast, in the boosting algorithm, base members are trained sequentially. According to \cite{breiman2001random}, due to the randomization technique used to curtail the correlation between individual trees' predictions, RF dominates other ensembles. Bootstrap aggregation (also known as bagging algorithm) uses the bootstrap strategy in which data are randomly alternated from the main data; the algorithm members are then trained on random subsamples of the original dataset \cite{breiman1996bagging}. However, a recent study \cite{lakshminarayanan2016simple} showed that in an ensemble of DNNs, when the base learner has several local optima, using bootstrap deteriorates the performance since the trained base learner observes just 63\% of unique data points.
As a result, here, the ensemble encompasses M networks that are trained independently on a similar version of the dataset. All networks have the same number of layers, but the number of neurons within each layer is random. While all networks are expected to behave similarly in areas with enough training data, inconsistent probabilities will be obtained where there are no data available. An empirical distribution generated from all the predictions of all networks can provide the single mean (posterior) and variance of one test data point. PE can also be defined as (\ref{eq:4})  \cite{van2020simple}:

\begin{equation}
{\hat{p}(y\vert x) \approx \frac{1}{M} \sum_{i=0}^{M} {p_{\theta_i}}(y\vert x)
\label{eq:3}}
\end{equation}

\begin{equation}
{PE = \sum_{i=0}^{C} \hat{p}({y_i}\vert x) \log \hat{p}({y_i}\vert x)
\label{eq:4}}
\end{equation}

\noindent where ${\theta_i}$ denotes the set of parameters of ${i_{th}}$ network element and $C$ denotes the classes. The low $PE$ value indicates that the prediction from all base learners is similar.

\subsection{Ensemble Monte Carlo dropout (EMCD)}
The EMCD algorithm is developed from the interleaving of an ensemble of deep networks and the MCD algorithm. The notion behind this model comes from a paper stating that combining the existing UQ techniques yields a new UQ method, which may potentially enhance the model's ability to capture uncertainty \cite{kendall2017uncertainties}. Here, ensemble is a combination of DNNs with various architectures and MCD carries out several stochastic forward passes to evaluate this ensemble of networks. Through the use of dropout and T stochastic forward passes at test time, each network generates multiple probability values for a given input. All probabilities will be represented as a Gaussian mixture distribution, which can be used to estimate both single mean (posterior) and variance. PE is applied the same as the ensemble algorithm, but the posterior is calculated differently (\ref{eq:5} and \ref{eq:6}) .

\begin{equation}
{\hat{p}(y\vert x) \approx \frac{1}{T} p(\hat{y}\vert \hat{x}, \hat{w})
\label{eq:5}}
\end{equation}

\begin{equation}
{PE = \sum_{i=0}^{C} \hat{p}({y_i}\vert x) \log \hat{p}({y_i}\vert x)
\label{eq:6}}
\end{equation}

\noindent where  $\hat{w_t}$   denotes the model parameters and $C$ denotes the number of classes.

   \begin{figure}[!t]
      \centering
      \includegraphics[scale=0.18]{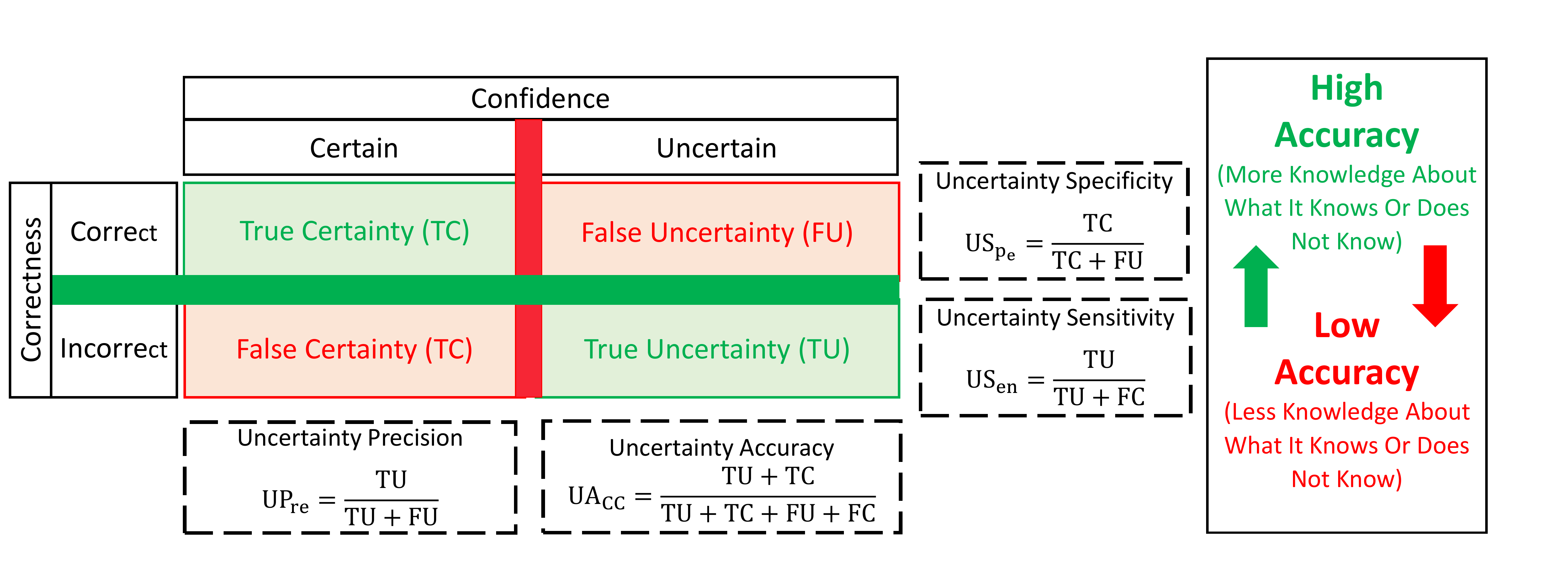}
      \caption{The UQ confusion matrix and performance metrics}
      \label{fig:one}
   \end{figure}

\section{Predictive uncertainty evaluation}
\label{sec:IV}
The confusion matrix is a beneficial tool to illustrate and compare the results of supervised learning models used for the classification task. This matrix shows the classifier performance on a given test data. Generally, the predicted test data are separated into two categories, correct and incorrect. By assuming the availability of the ground truth of the test data for a binary classifier, there are four sets of possibilities as follows: True positive $(TP)$, which is predicted as a correct label by the classifier, and its ground truth is labeled as correct. True negative $(TN)$, which is predicted as an incorrect label by the classifier, and its ground truth is also labeled as incorrect. False positive $(FP)$, which is predicted as a correct label by the classifier, but its ground truth is labeled as incorrect. False negative $(FN)$, which is predicted as an incorrect label by the classifier, but its ground truth is labeled as correct. Both $TP$ and $TN$ are the ideal outcomes. The weakness of the classifier is conspicuous through $FP$ and $FN$. Additionally, to evaluate the predictive uncertainty estimates an idea such as the confusion matrix was proposed \cite{asgharnezhad2020objective}. A threshold is used to cast predictions into certain and uncertain categories. Combining the two groups, the results are presented in Fig. \ref{fig:one}. The four possible combinations are true certainty $(TC)$ where the outcome is correct with certainty; true uncertainty $(TU)$ where the outcome is incorrect with uncertainty; false uncertainty $(FU)$ where the outcome is correct with uncertainty; and false certainty $(FC)$ where the outcome is incorrect with certainty. Both $TC$ and $TU$ are ideal results and correspond to $TN$ and $TP$, respectively. While $FU$ is almost favorable, $FC$ is the worst case. To quantify predictive uncertainty estimates, performance metrics, similar to the confusion matrix metrics, were also defined as below:

\begin{itemize}
\item Uncertainty sensitivity $(USen)$:  Defined as (\ref{eq:7}), $USen$ corresponds to the ratio of incorrect predictions that are flagged as uncertain predictions to all the incorrect predictions (certain or uncertain). Sensitivity, also known as recall, in the traditional confusion matrix is defined as $TP⁄(TP+FN)$, which is the ratio of $TPs$ to all positives. High recall indicates that very few positives were incorrectly classified as negatives. High uncertainty recall implies that very few incorrect predictions were erroneously flagged as certain predictions. 

\begin{equation}
{USen = \frac{TU}{TU+FC}
\label{eq:7}}
\end{equation}

\item Uncertainty specificity $(USpe)$: $USpe$ is defined as (\ref{eq:8}). This measure corresponds to the ratio of correct predictions that are flagged as certain predictions to all the correct predictions whether certain or uncertain. Specificity in conventional confusion matrix is defined as $TN⁄(TN+FP)$, which is the proportion of $TNs$ to all negatives. High specificity indicates that very few negatives were incorrectly classified as positives. High $USpe$ implies that very few correct predictions were erroneously flagged as uncertain predictions. 

\begin{equation}
{USpe = \frac{TC}{TC+FU}
\label{eq:8}}
\end{equation}

\item Uncertainty precision $(UPre)$: $UPre$ is defined as (\ref{eq:9}) and measure the ratio of incorrect predictions that are flagged as uncertain predictions to all the predictions with uncertainty whether correct or incorrect. Precision in conventional confusion matrix is defined as $TP⁄(TP+FP)$, which is the ratio of $TPs$ to the total predicted positive samples. High precision implies very few negatives were incorrectly classified as positives. High $UPre$ indicates that very few correct predictions were erroneously flagged as uncertain predictions.

\begin{equation}
{UPre = \frac{TU}{TU+FU}
\label{eq:9}}
\end{equation}

\item Uncertainty accuracy $(UAcc)$: $UAcc$ is defined as (\ref{eq:10}). It corresponds to all the values lying across the “main diagonal” divided by the total number of outcomes. Accuracy, which is used in studying normal classifications, is defined as $(TP+TN)⁄(TP+TN+FP+FN)$.

\begin{equation}
{UAcc = \frac{TU+TC}{TU+TC+FU+FC}
\label{eq:10}}
\end{equation}

\end{itemize}
Ideally, recall, specificity, precision, and accuracy should all be close to 1 in the traditional confusion matrix. The $FN$ rate and $FP$ rate should be close to 0. As illustrated in Fig. \ref{fig:one}, a favorable model should have high uncertainty accuracy. The best result for the uncertainty metrics is 1 and the worst is 0. Having the best result in the above metrics signifies that the model is conscious of what it knows and what it does not know. It clarifies when a prediction is reliable and when it is not.

\begin{figure}[!t]
      \centering
      \includegraphics[width=\columnwidth]{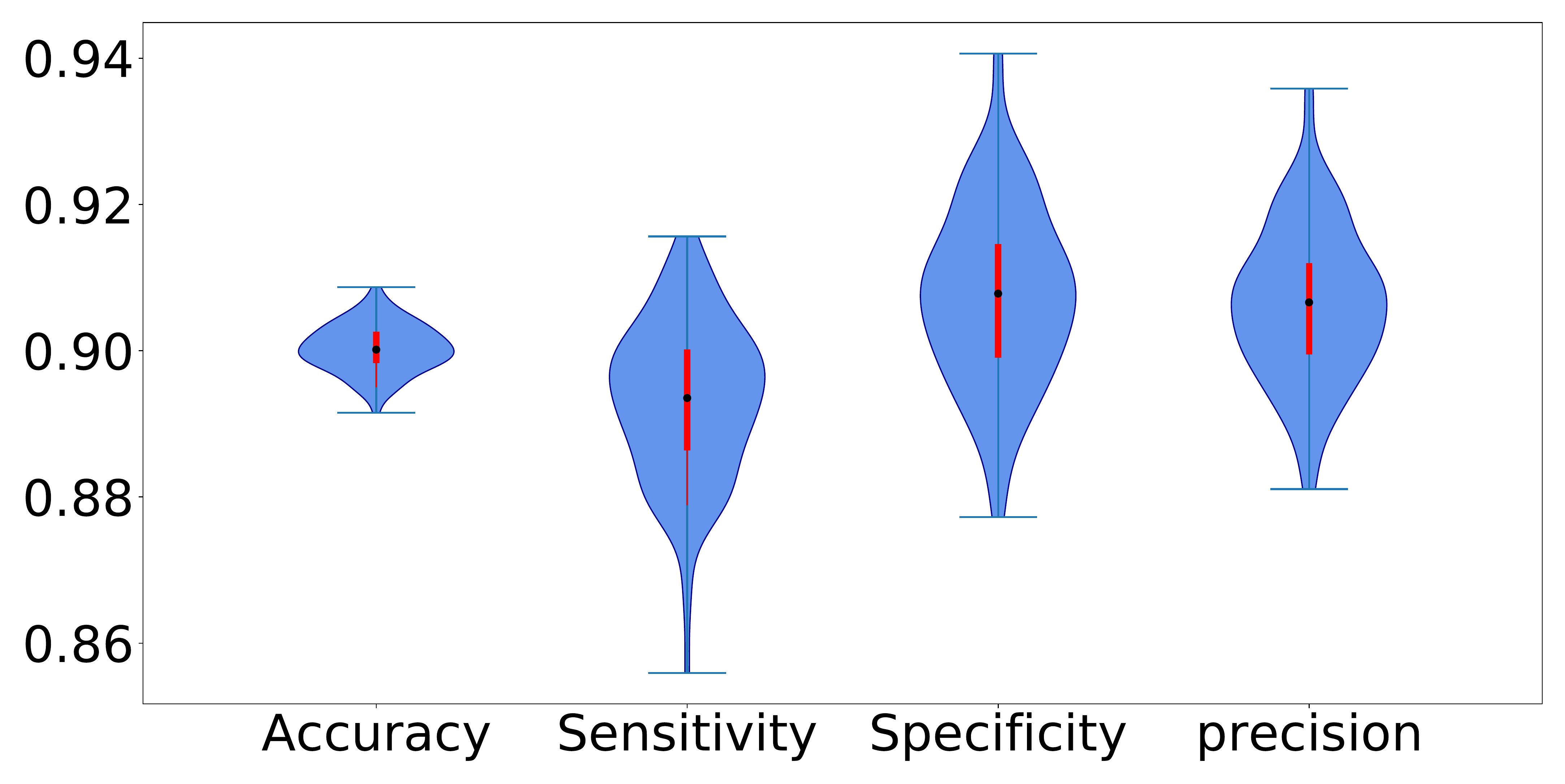}
      \caption{Violin plots depict the distribution of accuracy, sensitivity, specificity and precision performance metrics of a single and simple neural network model after 100 evaluations using transaction data.}
     \label{fig:two}
   \end{figure}

\section{Experimental Setup}
\label{sec:V}

\subsection{Dataset}
This study is conducted using a publicly available dataset provided by the Vesta Corporation \cite{kaggle}. This high-dimensional dataset encompasses 385 features and 41,326 records. There are both categorical and numerical variables within the existing data. An overview of some of the variables that embrace the details of transaction and customer ID includes card numbers, type of card (debit, credit, and charge), type of device, amount of transaction, product/service code, the interval between transactions, email address, numerous engineered features of the transaction, etc. Due to privacy concerns, customers are identified by numbers rather than their actual names. In total, 20663 feature vectors belong to fraudulent transactions, which is the same number as the non-fraudulent ones. Thus, the proportion of the two classes is relatively balanced.

\subsection{Experiments}
In practice, the dataset has many missing values that could undermine the algorithm efficiency. To address this issue, we employ the simple-imputer technique to impute these values. We operate the standard-scaler to make a similar scale for numerical variables and ordinally encod the categorical features. Before processing, the whole dataset is divided into training and test sets (70\% and 30\%, respectively). The fully connected neural network is comprised of an input layer with 385 units, three hidden layers, and an output layer. The softmax function is also assigned top of the pipeline. The rectified linear activation function (Relu), dropout rate of 0.30, and 50 epochs are exerted for model development. The adaptive learning rate method (Adam) with a learning rate of 0.001 is utilized to optimize the cross-entropy loss function. For the MCD model, the number of neurons in each layer is 256, 64, and 16, respectively. For the ensemble model, 30 networks are created, all of which have three hidden layers, but the number of neurons in each layer is picked at random between (256, 385), (64, 256), and (16, 32), respectively. The EMCD algorithm is the same as the ensemble, except that the MCD algorithm is used to evaluate each network (see Table \ref{tab:Table1}).

\begin{table}
\centering
\caption{Configurations of proposed methods}
\begin{tabular}{l|cccc} 
\toprule
\multicolumn{2}{l}{Method}                                  & MCD     & Ensemble   & EMCD        \\ 
\toprule
\multicolumn{2}{l}{Units of the input layer}                & 385     & 385        & 385         \\ 
\cmidrule(lr){1-5}
\multirow{3}{*}{\rotcell{Units of hidden layers}} & Layer 1 & 256     & (256, 385) & (256, 385)  \\ 
\cmidrule(lr){2-5}
                                                  & Layer 2 & 64      & (64, 256)  & (64, 256)   \\ 
\cmidrule(r){2-5}
                                                  & Layer 3 & 16      & (16, 32)   & (16, 32)    \\ 
\cmidrule(lr){1-5}
\multicolumn{2}{l}{Output layer activation function}        & Softmax & Softmax    & Softmax     \\ 
\cmidrule(r){1-5}
\multicolumn{2}{l}{Dropout rate}                            & 0.3     & 0.3        & 0.3         \\ 
\cmidrule(lr){1-5}
\multicolumn{2}{l}{Activation function}                     & Relu    & Relu       & Relu        \\ 
\cmidrule(lr){1-5}
\multicolumn{2}{l}{Adam learning rate}                      & 0.001   & 0.001      & 0.001       \\
\toprule
\end{tabular}
\label{tab:Table1}
\end{table}

\section{Discussions and results}
\label{sec:VI}
First, we examine how well a single and simple DNN model does in classification. For this purpose, the best configuration of one model among 30 trained models in an ensemble is used as the basis of a new model. Moreover, it is reasonable to divide the whole dataset randomly into training and test sets with the same proportions of samples for each class. In Fig. \ref{fig:two}, the violin plots of the accuracy, sensitivity, specificity and precision metrics of this model after 100 evaluations are shown. With a standard deviation of 0.003, the average of all accuracies is 0.90 and it appears that the model is qualified to trap suspicious transactions as fraud.
\begin{figure*}
  \centering
  \subfloat[][MCD]{\includegraphics[width=.32\textwidth]{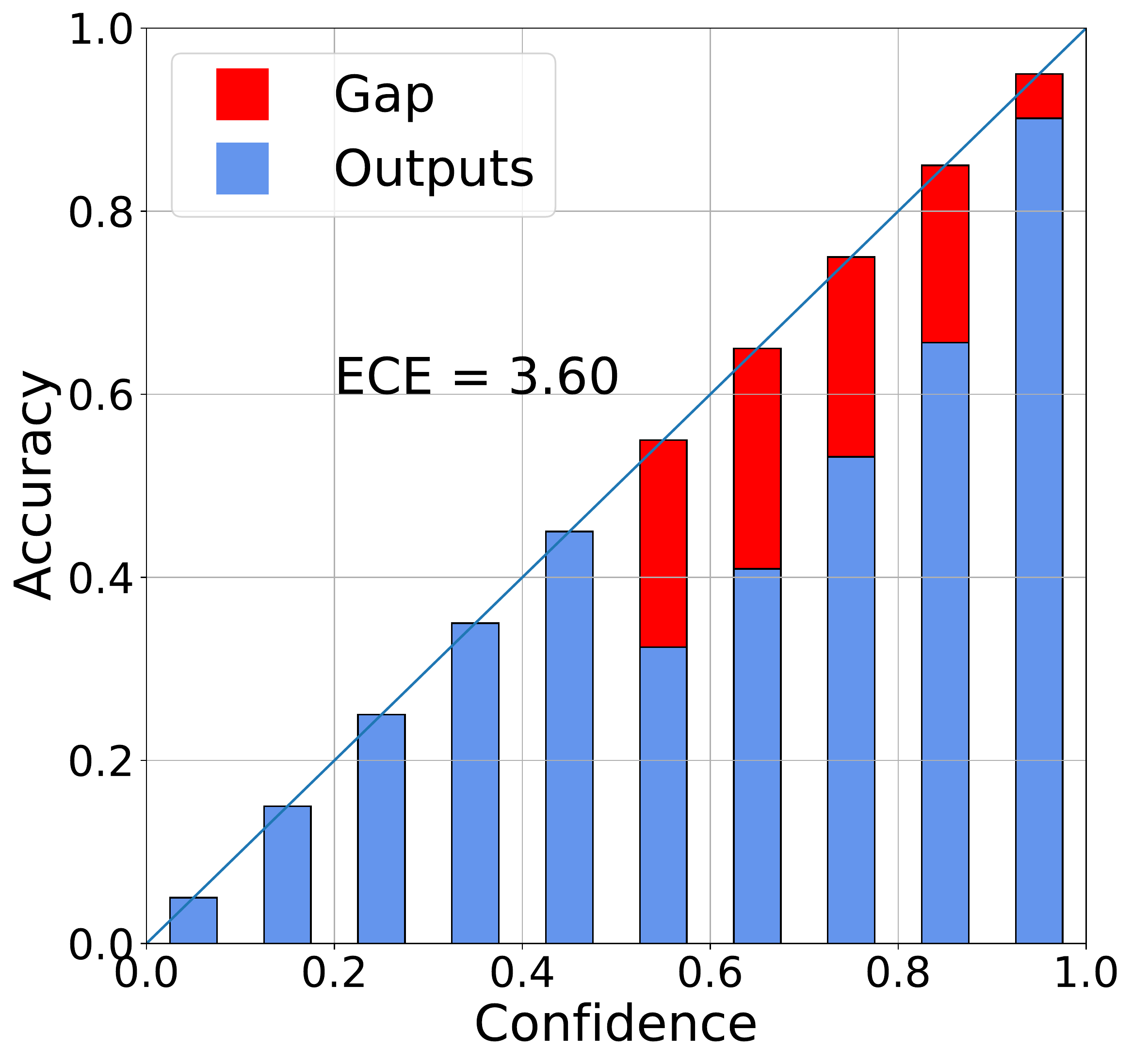}}
  \subfloat[][Ensemble]{\includegraphics[width=.32\textwidth]{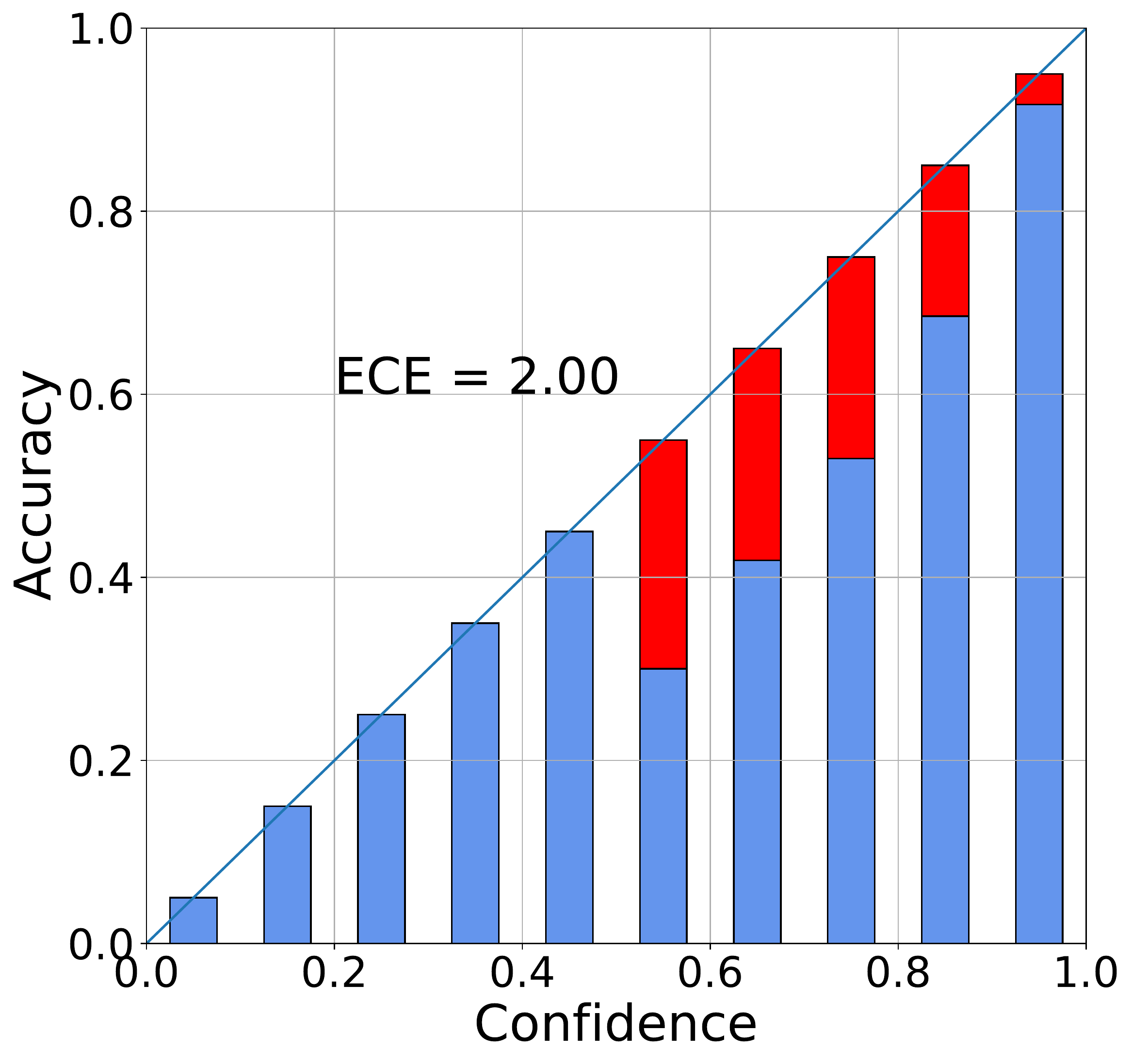}}
  \subfloat[][EMCD]{\includegraphics[width=.32\textwidth]{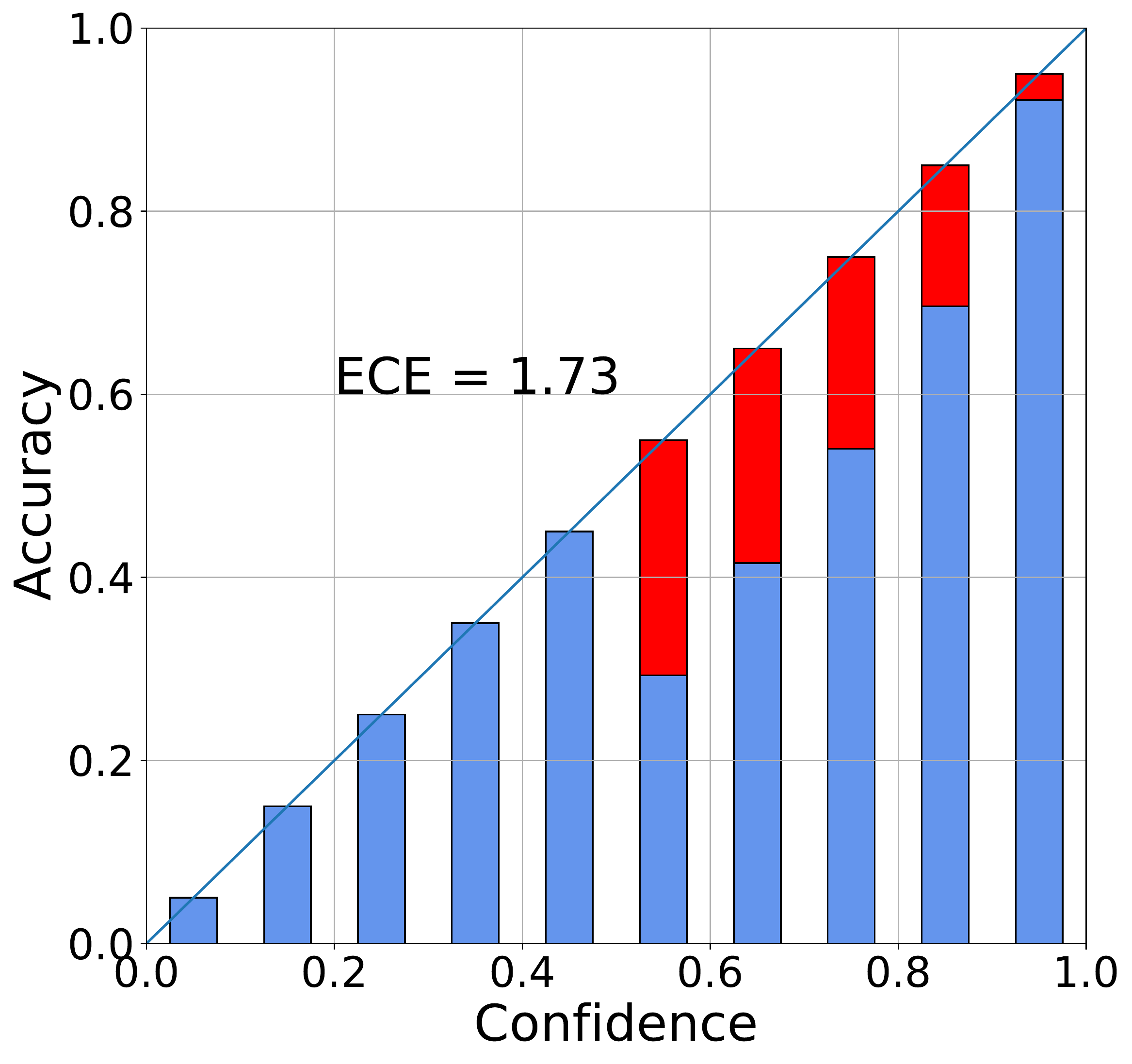}}\par
  \caption{The reliability diagrams of the trained DNNs. The red section is called the gap, and a smaller gap is better. High ECE values obtained by three models reveal the miscalibration of generated probabilities. EMCD achieves the best calibration results amongs UQ methods.}
  \label{fig:three}
\end{figure*}

The DNN outputs are the probabilities that are the approximation scores of the predictions (confidence) along with the corresponding labels extracted by the model. The first definition of results obtained with the single model only provides a row of classification results, not confidence results. Modern DNNs are prone to produce uncalibrated results, meaning that compared with expected accuracy, the probability is overestimated. Confidence in classifiers is measured using calibration, which indicates how well the probabilities predicted for each class (probability of correctness) align with the empirical accuracy. As stated in \cite{guo2017calibration}, the expected calibration error (ECE) groups the probability interval into a specified number of buckets and assigns each predicted probability to the bucket that encompasses it. The calibration error is the difference between the fraction of correct predictions in the bucket (accuracy) and the mean of the probabilities in the bucket (confidence). Thus, ECE is computed as the average gap between the within-bucket accuracy and the within-bucket predicted probability for M buckets (\ref{eq:11}).

\begin{figure*}
  \centering
  \subfloat[][MCD]{\includegraphics[width=.333\textwidth]{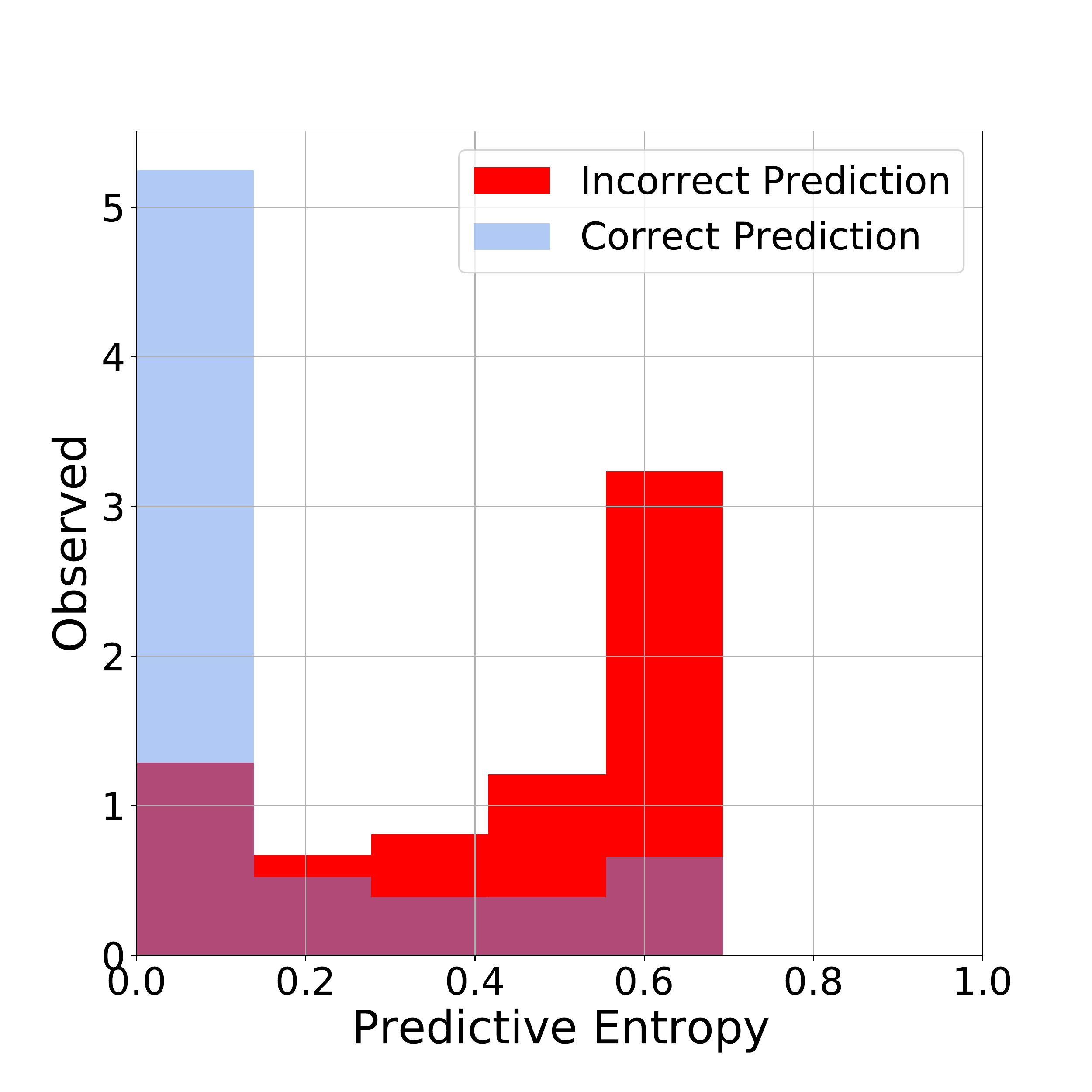}}\hfill
  \subfloat[][Ensemble]{\includegraphics[width=.333\textwidth]{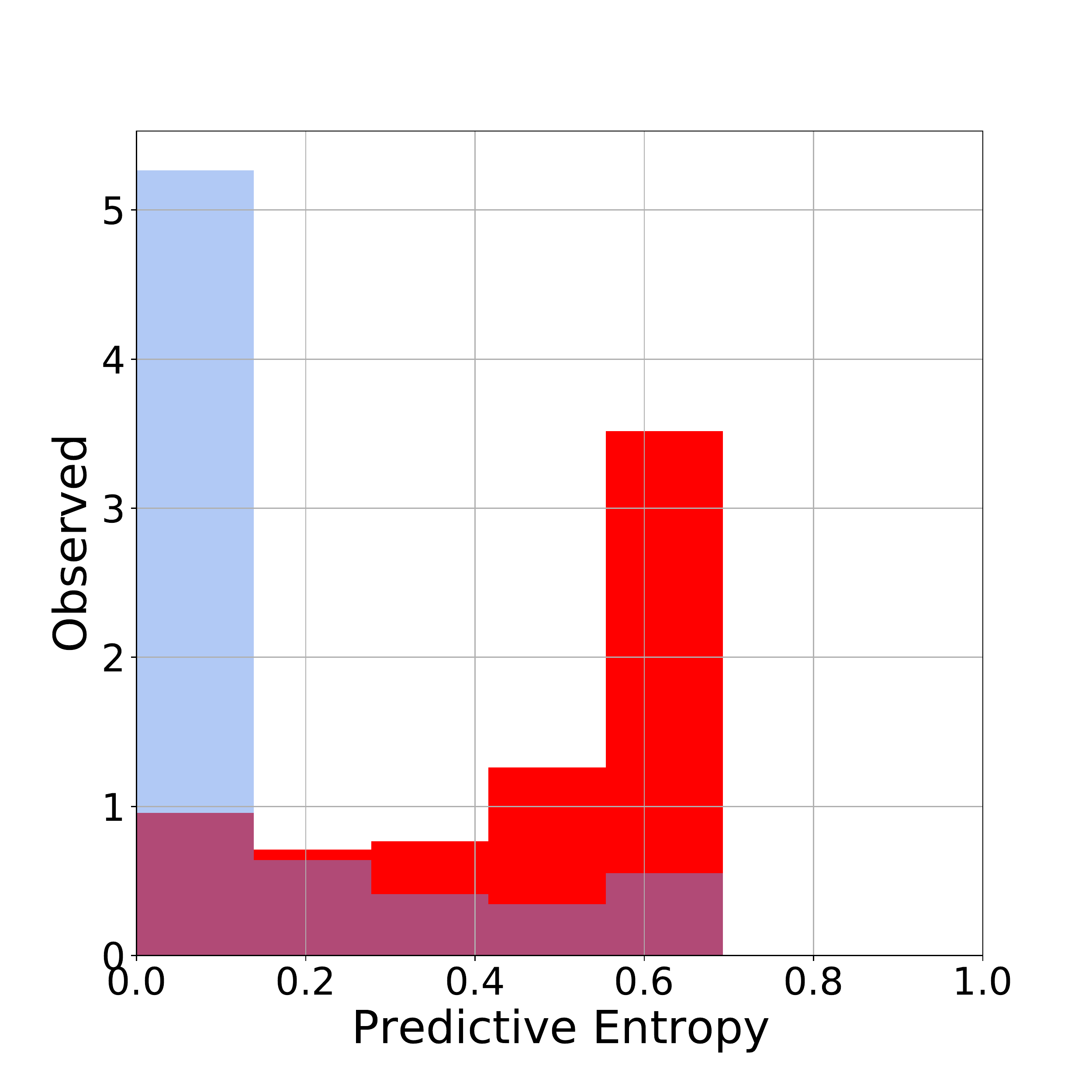}}\hfill
  \subfloat[][EMCD]{\includegraphics[width=.333\textwidth]{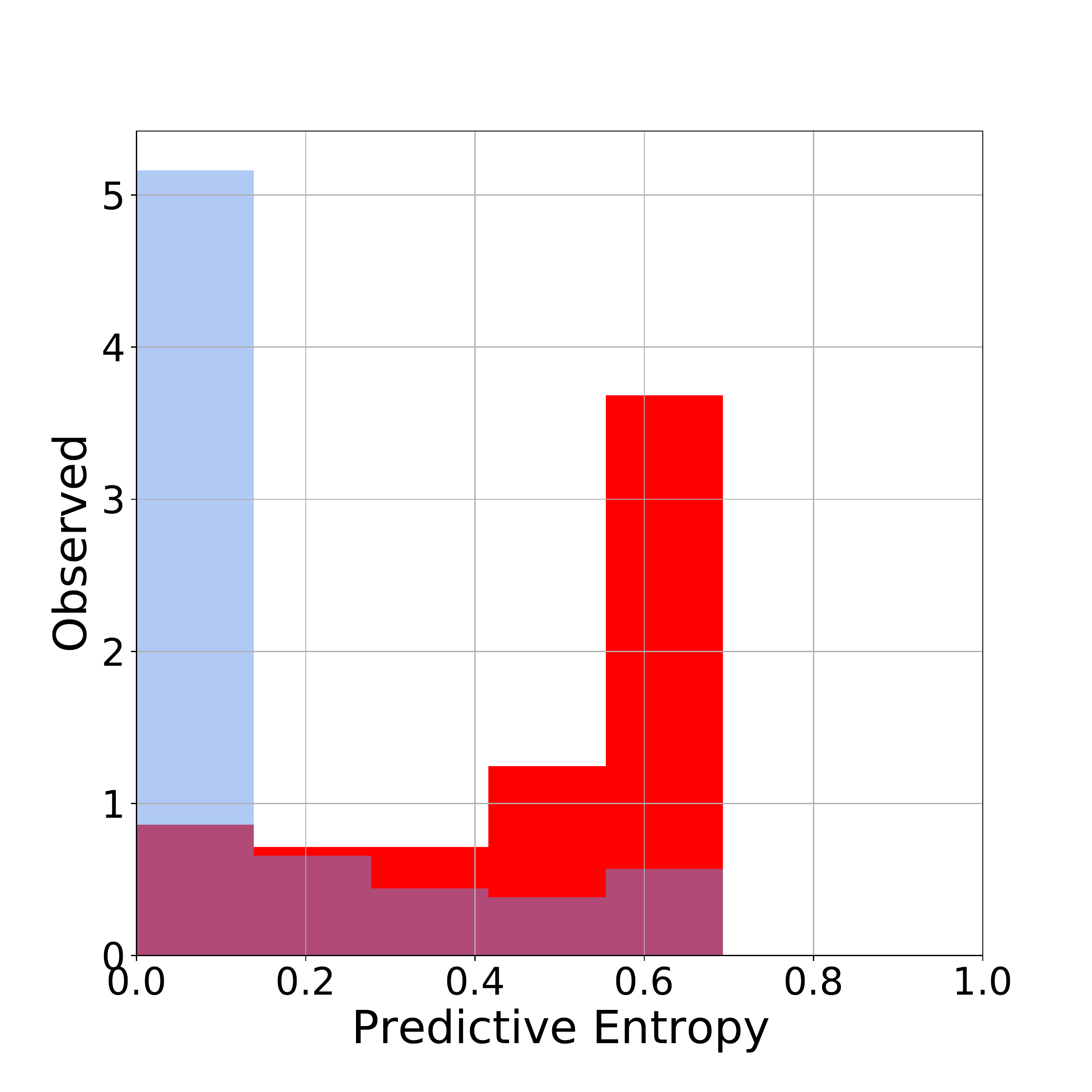}}\par
  \caption{Histogram graphs indicate the predictive entropy results for three UQ methods using feature vectors. Considering the ensemble method, most of the correct predictions are captured with a high certainty level (entropy close to zero) as there is an agreement among ensemble members regarding the predicted labels. In contrast, most of the incorrect predictions are flagged with a high uncertainty level (entropy close to maximum) as there is an inconsistency between ensemble members regarding the predicted labels. EMCD and MCD, irrespective of their procedural difference, obey a similar approach.}
  \label{fig:four}
\end{figure*}

\begin{equation}
{ECE = \sum_{m=1}^{M} \frac{\vert {B_m}\vert }{n} \vert Acc({B_m}) - Conf({B_m})\vert
\label{eq:11}}
\end{equation}

\begin{equation}
{Acc({B_m}) = \sum_{m=1}^{M} \frac{1}{\vert {B_m} \vert} 1(\hat{{y_i}} = {y_i})
\label{eq:12}}
\end{equation}

\begin{equation}
{Conf({B_m}) = \sum_{m=1}^{M} \frac{1}{\vert {B_m} \vert}
\label{eq:13}}
\end{equation}

\noindent where ${B_m}$  is the number of predictions in bucket $m$, $n$ is the total number of data points, $Acc({B_m})$ (\ref{eq:12}) and $Conf({B_m})$ (\ref{eq:13}) are the accuracy and confidence of bucket $m$, respectively. Also \textbf{1}$($·$)$ is the indicator function. Hence, before commencing the uncertainty analysis, we first verify the calibration of predictions generated by the three proposed DNN models through the ECE diagram. Based on the ECE notion, the ECE diagrams are plotted in Fig. \ref{fig:three} as a visual tool for evaluating the calibration. These diagrams plot expected sample accuracy as a function of confidence. In Fig. \ref{fig:three}, three reliability diagrams specify that none of the probabilities generated by the proposed DNN models are calibrated. This figure demonstrates that these classifiers reflect samples in an overconfident manner that could be detrimental. The ensemble and EMCD obtain lower ECE values. It is also imperative to note that the model can be calibrated through a variety of procedures, although this does not eliminate the reliance on UQ techniques. A well-calibrated model would obtain a low ECE leading to reliable and favorable results. Even results obtained by well-calibrated (low ECE) model with scant accuracy is more reliable.

We consider 1000 iterations of MC for both MCD and EMCD methods. In all the methods, the class with the bigger softmax output from the last activation function in the pipeline for the distribution mean is considered as the predicted outcome. Then the entropy equations (\ref{eq:2}), (\ref{eq:4}), and (\ref{eq:6}) are applied for each method to estimate the uncertainty associated with this outcome.

\begin{figure*}
  \centering
  \subfloat[][MCD]{\includegraphics[width=6cm, height=6cm]{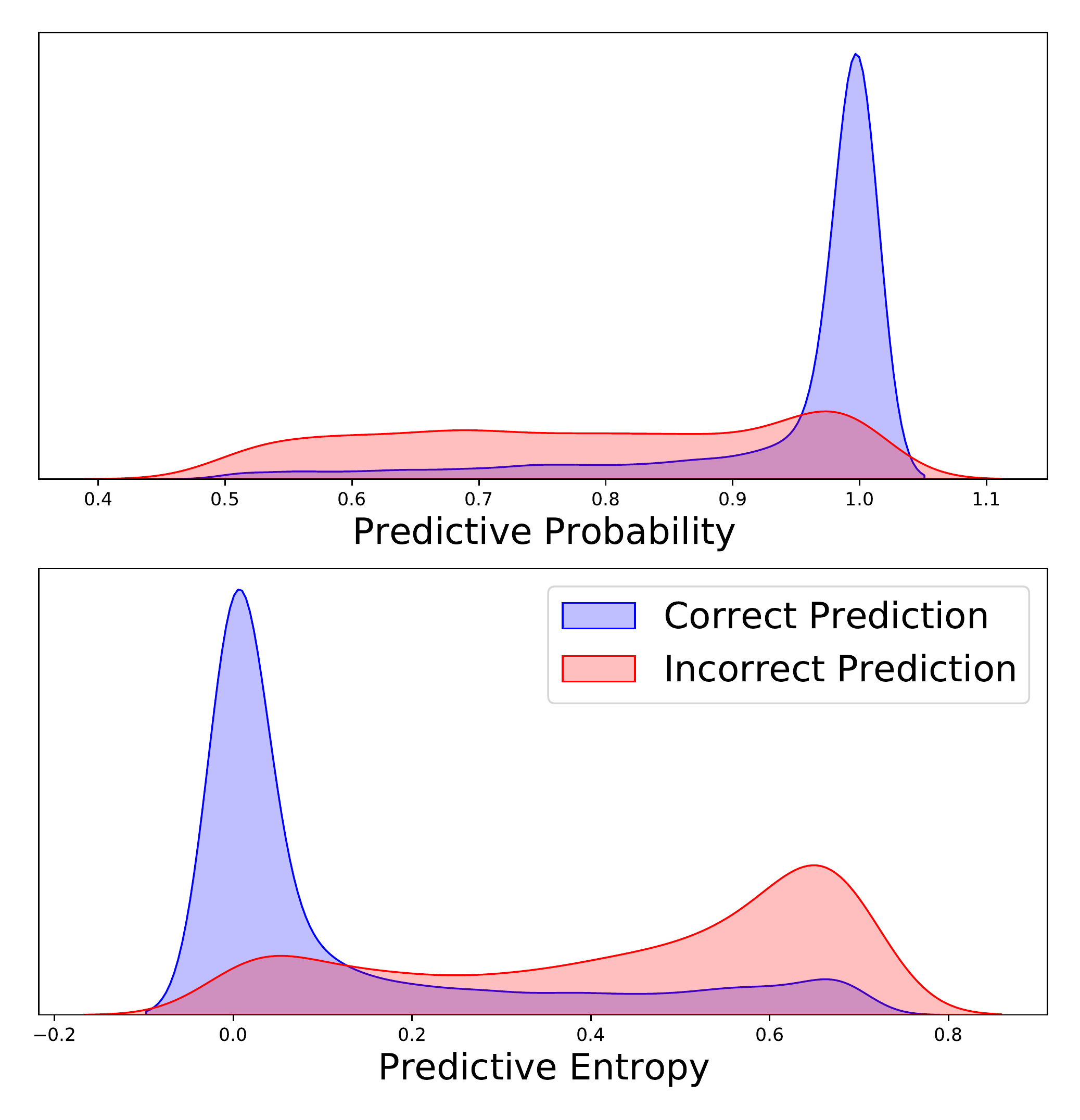}}\hfill
  \subfloat[][Ensemble]{\includegraphics[width=6cm, height=6cm]{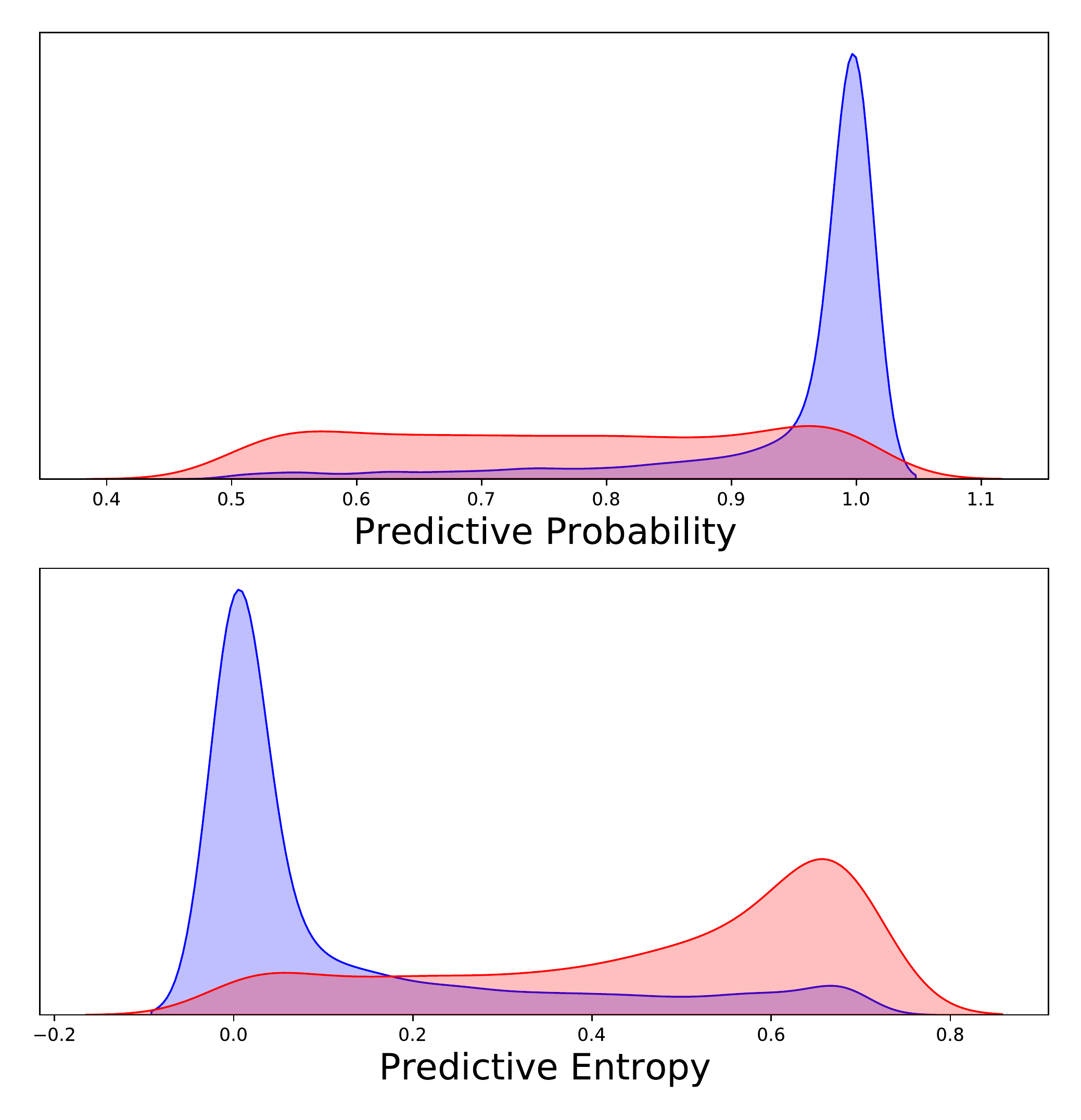}}\hfill
  \subfloat[][EMCD]{\includegraphics[width=6cm, height=6cm]{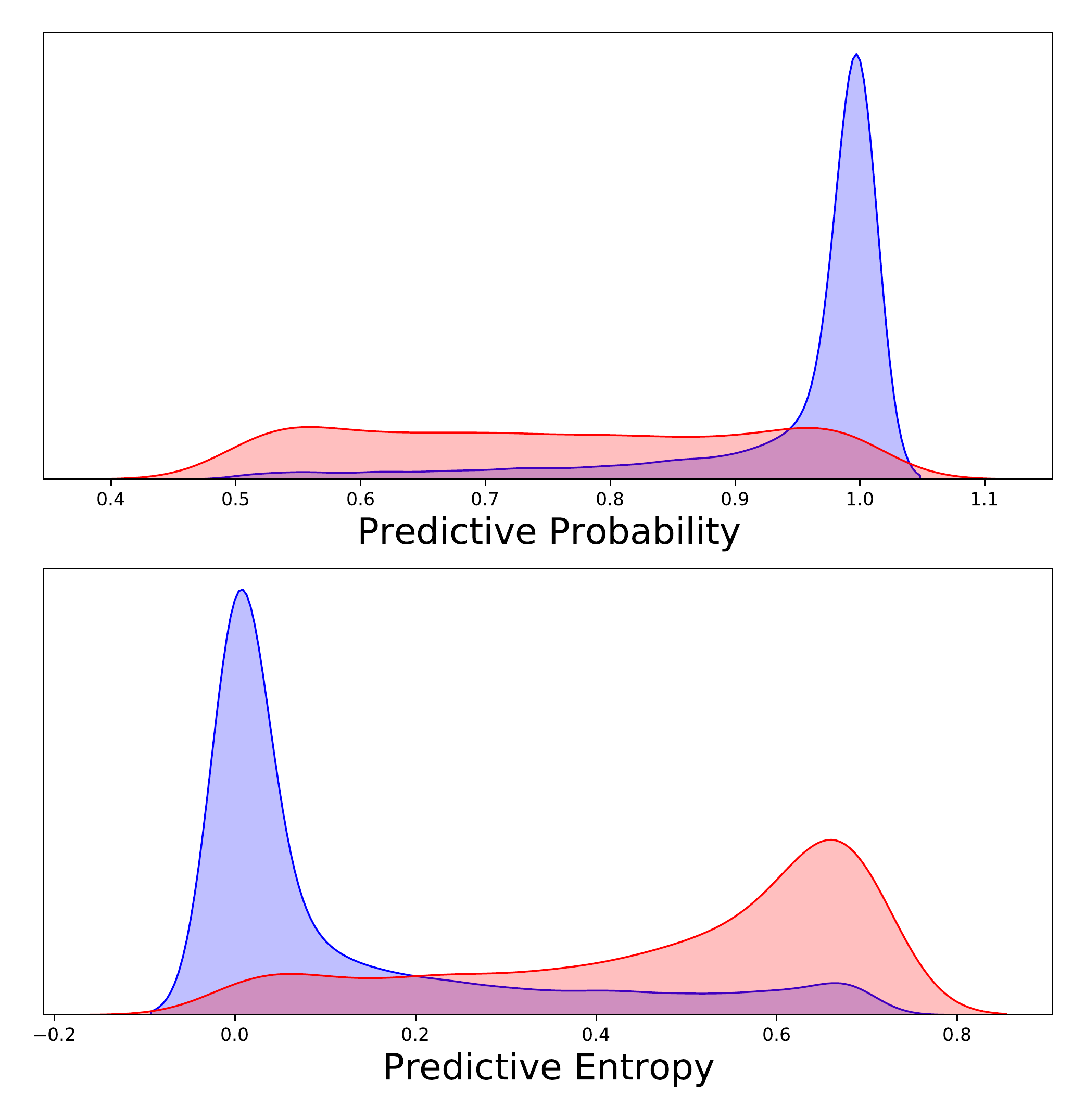}}\par
  \caption{The density plots show the distribution of the predicted probabilities and the distribution of estimated predictive uncertainty estimates (entropy) for three UQ methods. Correct classification and misclassification predictions are delineated in each plot. In terms of the best separation between two groups, there is no considerable difference between the UQ techniques (bottom row).}
  \label{fig:five}
\end{figure*}

The histogram diagrams in Fig. \ref{fig:four} demonstrate the predictive entropy (PE) of test data set for each proposed model. The obtained PE results after comparison with ground truth labels are categorized to correct and incorrect groups. In the context of MCD analysis, variation in predicting outcomes is associated with the lack of model confidence. The almost steady stream of outputs in every iteration of one test data point highlights the model’s confidence regarding the predicted sample. Ensemble and EMCD, regardless of their procedural difference, obey a similar approach. As depicted in the histogram diagram in Fig. \ref{fig:four}, highly certain correct predictions (TC) skew to the left of the histogram (blue peaks with entropy close to 0) while the uncertain incorrect predictions (TU) shift from left to the right of the histogram (red peaks with entropy close to 1). Certain incorrect predictions (FC) and uncertain correct predictions (FU), which manifest UQ model inconsistency, are located in the center of diagram (between two blue and red peaks). Fig. \ref{fig:four} reveals that the majority of incorrect predictions are associated with significant level of uncertainty. In other words, models have confessed \textit{“Do not trust my uncertain outputs”}, which gives users a chance to conduct the inspection to avoid possible losses. A model can be considered confident when incorrect predictions have entropy close to 1 (i.e. greater uncertainty) as well as correct predictions with entropy close to 0 (i.e. greater certainty). This means that in Fig. \ref{fig:four}, the two blue and red peaks should place at left and right of the diagram respectively with the maximum possible distance along with the maximum smoothness between these two peaks. Note that Fig. \ref{fig:four} provides qualitative quantification of uncertainty to reflect a general comparison between proposed models. Accordingly, it is not accurate to infer the finest model based on Fig. \ref{fig:four} results.

\begin{figure}
  \centering
  \subfloat[][UAcc]{\includegraphics[width=.5\columnwidth, height=4cm]{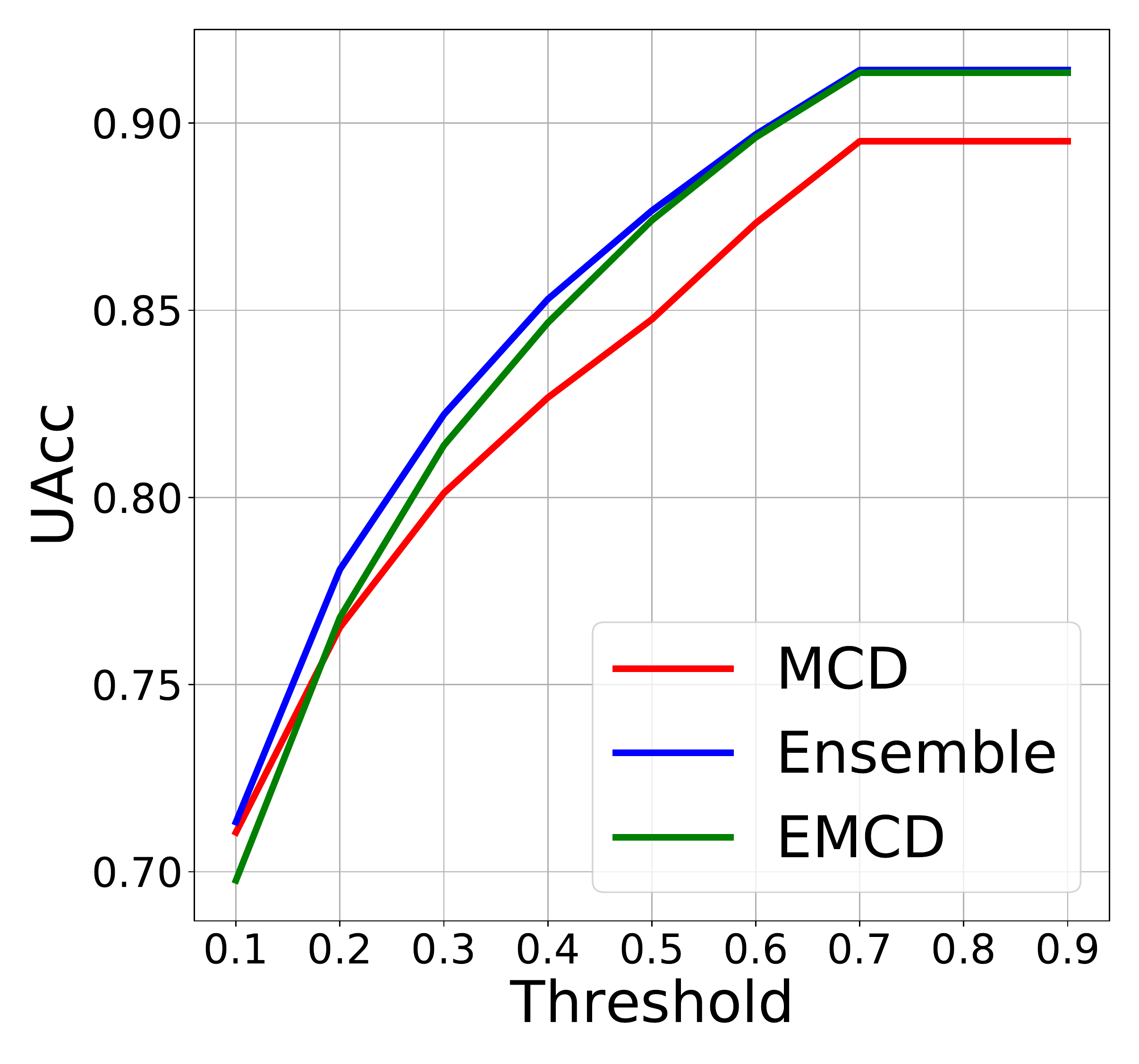}}\hfill
  \subfloat[][USen]{\includegraphics[width=.5\columnwidth, height=4cm]{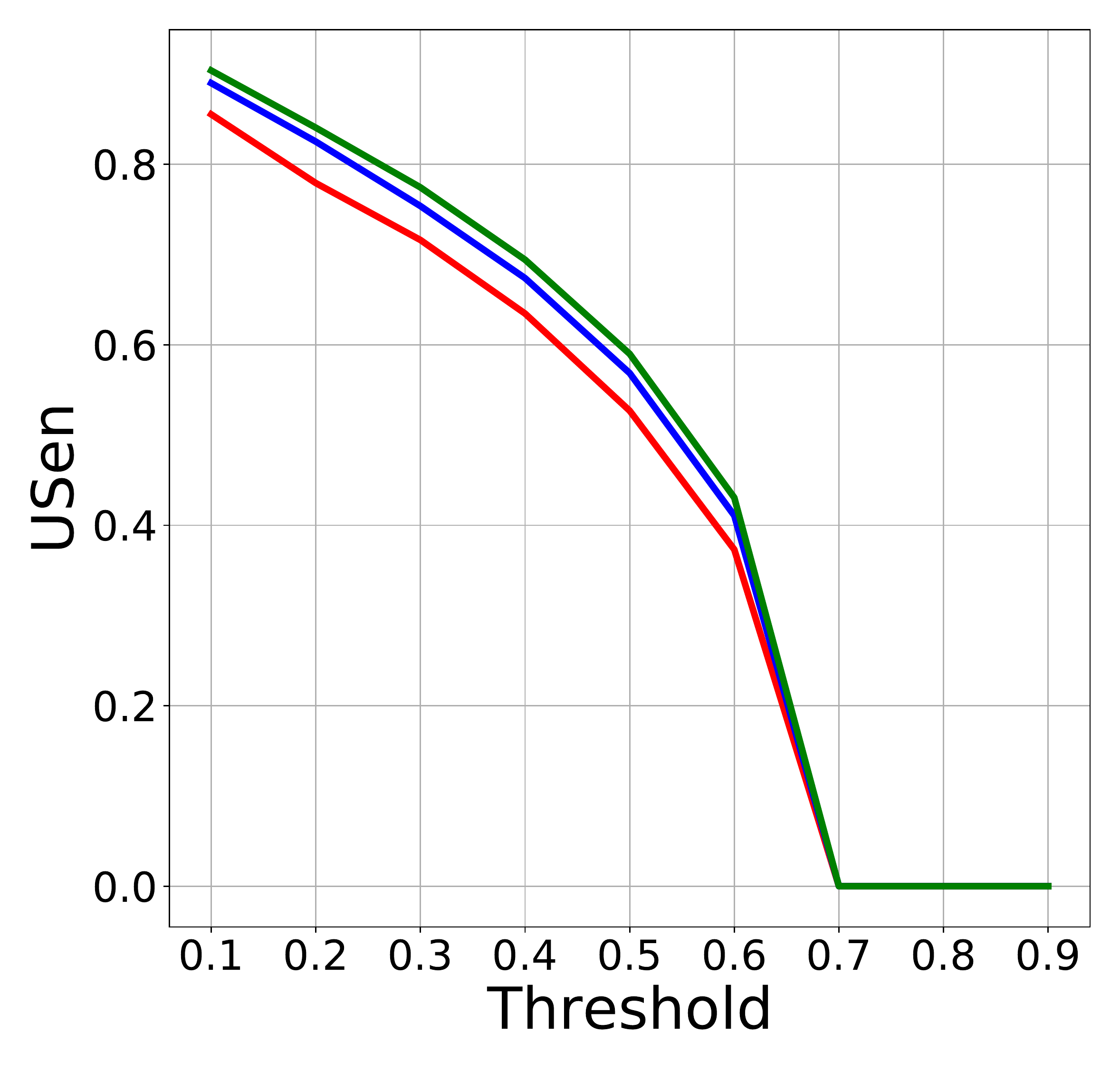}}\hfill
  \subfloat[][USpe]{\includegraphics[width=.5\columnwidth, height=4cm]{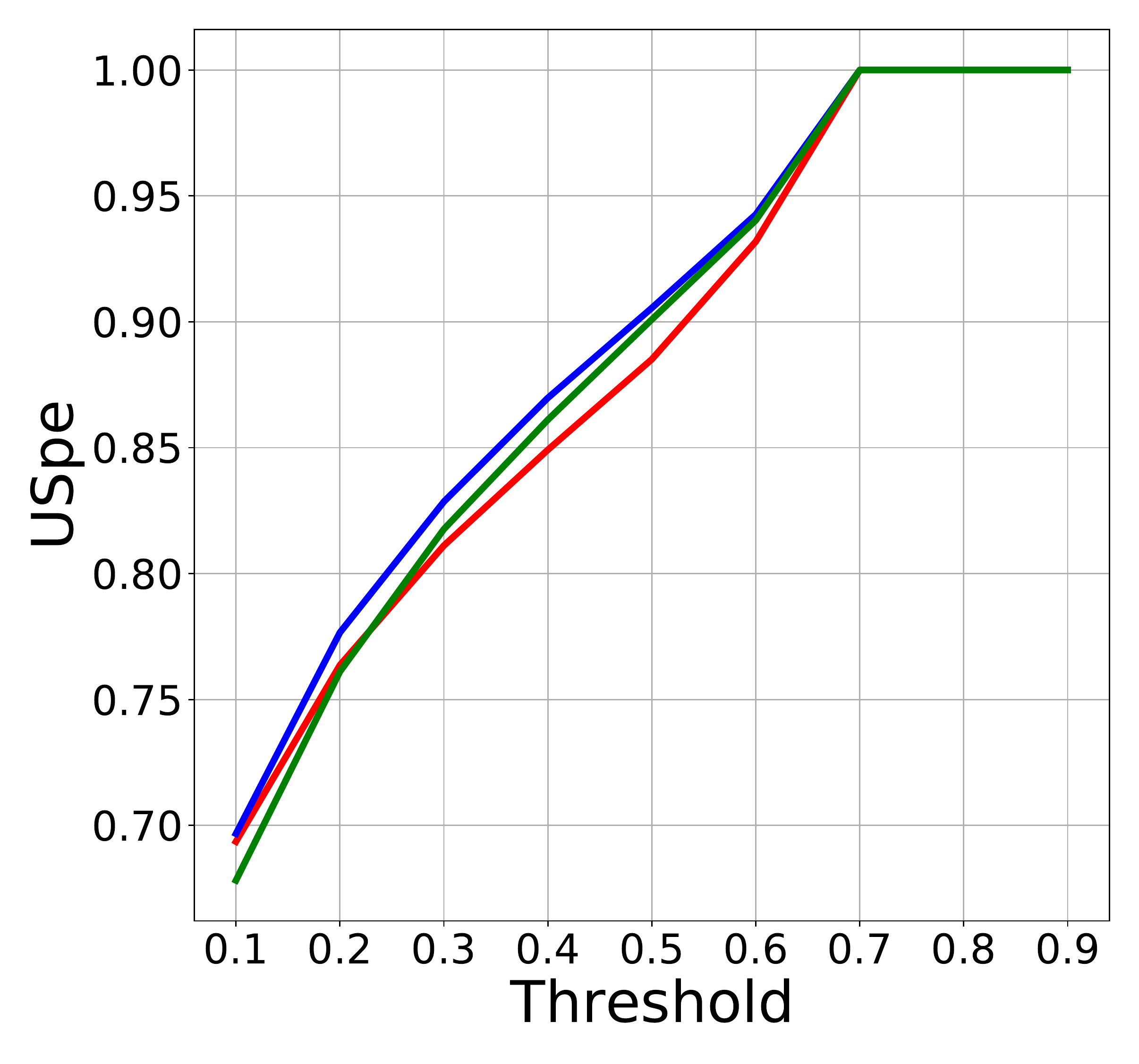}}\hfill
  \subfloat[][UPre]{\includegraphics[width=.5\columnwidth, height=4cm]{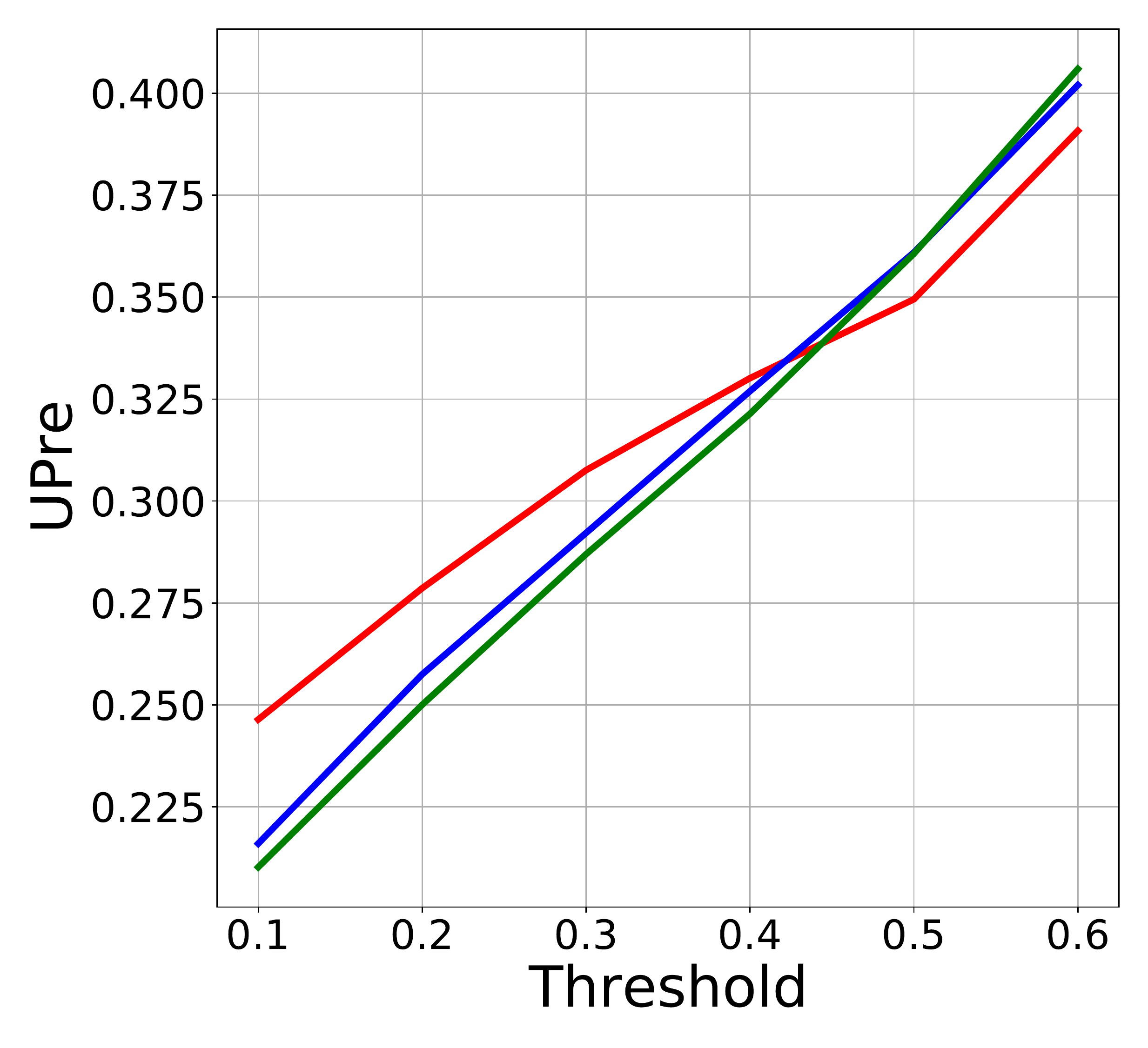}}\par
  \caption{Evaluating three UQ techniques using performance metrics (uncertainty accuracy, sensitivity, specificity, and precision) are plotted for threshold values in the range of 0.1 and 0.9.}
  \label{fig:six}
\end{figure}

In Fig. \ref{fig:five}, kernel density estimation (KDE) plots of predicted probabilities and entropy estimates, corresponding to the value of the max softmax output, for three UQ methods are illustrated. The correct classification and the misclassification predictions are well grouped in each plot. Due to the high accuracy of the models in the accurate predictions, the centers of the two groups are relatively apart from each other. The correct classified group tends to have a more centered distribution, whereas the incorrect group has a more scattered distribution. Meanwhile, the estimated uncertainties are bigger for misclassified feature vectors. As it can be observed from the three predictive entropy plots, similar to the results in Fig. \ref{fig:four}, the mean of the uncertainty of misclassified group is greater than the correct classified one. The findings of plots clarify the model promising ability to express its confidence (or lack thereof) in generated predictions. The results are practically important since reliable uncertainty estimates provide much-needed insight into the predicted probabilities. Investigators can use these techniques to flag uncertain predictions as suspicious transactions.
\\\indent Additionally, the performance metrics explained in section \ref{sec:IV} help to evaluate the predictive uncertainty estimates. In Fig. \ref{fig:six}, the UAcc, USen, USpe, and UPre metrics are illustrated. The value of these metrics depends on a threshold which defines the boundary between certainty and uncertainty. By changing the threshold, values of TC, FC, TU and FU would change leading to variant in aforementioned metrics. In this study, further analysis is done by setting the thresholds between 0.1 and 0.9. In contrast to USen, as the rate of threshold increases, the rates of UAcc, USpe, and UPre increase as well. There is a little difference between these three metrics in terms of direction at variant thresholds. High USen implies that all the three UQ methods can flag incorrect predictions and trigger an uncertainty report to appear. Since many correct predictions are uncertain (FU), UQ methods fail to attain a high UPre. Indeed, the number of feature vectors that are correctly classified is higher than the number of misclassified feature vectors, which results in a low UPre.
\\\indent Setting the threshold value depends on how important or sensitive the issue is to the stakeholders or end-users. Preferably, we consider 0.4 as the threshold rate for the three UQ techniques. Based on the empirical data, the uncertainty performance metrics for all the methods are compared in Table \ref{tab:Table2}. While the deep ensemble networks achieve 0.85 UAcc, this result for MCD and EMCD is 0.82 and 0.84, respectively. This corroborates that the ensemble model is somewhat superior. 

\begin{table}
\centering
\caption{Comparison of uncertainty performance metrics for three UQ methods using 12,398 feature vectors of transactions (threshold set to 0.4)}
\resizebox{\columnwidth}{!}{%
\begin{tabular}{lcccc}
\hline
{\tiny UQ Method} & {\tiny UAcc} & {\tiny USen} & {\tiny USpe} & {\tiny UPre} \\
\hline
{\tiny MCD} & {\tiny 0.82} & {\tiny 0.63} & {\tiny 0.84} & {\tiny 0.33} \\
{\tiny Ensemble} & {\tiny 0.85} & {\tiny 0.67} & {\tiny 0.86} &  {\tiny 0.32}\\
{\tiny EMCD} & {\tiny 0.84} & {\tiny 0.69} & {\tiny 0.86} & {\tiny 0.32} \\
\hline
\end{tabular}%
}
\label{tab:Table2}
\end{table}

\section{Conclusion and Future Research}
\label{sec:VII}
International e-commerce is rapidly expanding throughout the world, which has led to an increase in card fraud resulting in millions of dollars lost every year. In many existing studies and practical applications, DNNs have been broadly applied to detect card fraud. These networks fail to reliably measure and report their predictive confidence. In this study, three deep UQ techniques (MCD, ensemble, and EMCD) are introduced to measure the degree of uncertainty associated with generated predictions, which can lead to a reliable classification. In fraud detection, capturing uncertainty of predicted fraud transactions, provide extra insights for decision makers to enhance fraud prevention process. UQ confusion matrix and several performance metrics are also used to evaluate the predictive uncertainty estimates. Following empirical findings, we observe that the ensemble method is more effective in capturing uncertainties corresponding to predictions. \\ For future research in the realm of card fraud detection, these proposed UQ models can be extended or ameliorated in several ways. A further possibility seems to be the concrete dropout technique. In this case, the optimal rate of dropout in each layer is evaluated, leading to a well-calibrated estimate of UQ in the MCD and EMCD methods. Another approach is to optimally combine uncertainty estimates from three UQ methods to improve the quality of final uncertainty estimates.  



\end{document}